\documentclass[sigconf,authorversion,nonacm]{acmart}

\usepackage{algorithm}
\usepackage{algorithmic}
\usepackage{diagbox}
\usepackage{multirow, makecell}
\usepackage{amsmath}
\usepackage{multicol}
\usepackage{tablefootnote}
\usepackage{enumitem}
\usepackage{subfigure}

\AtBeginDocument{%
  \providecommand\BibTeX{{%
    \normalfont B\kern-0.5em{\scshape i\kern-0.25em b}\kern-0.8em\TeX}}}

\begin{document}

\title{Fairness for Text Classification Tasks with \\ Identity Information Data Augmentation Methods}

\author{Mohit Wadhwa, Mohan Bhambhani, Ashvini Jindal, Uma Sawant, Ramanujam Madhavan}
\affiliation{LinkedIn Corporation}
\email{(mwadhwa, mbhambhani, ajindal, usawant, rmadhavan)@linkedin.com}


\begin{abstract}
  Counterfactual fairness methods address the question: \textit{How would the prediction change if the sensitive identity attributes referenced in the text instance were different?} These methods are entirely based on generating counterfactuals for the given training and test set instances. Counterfactual instances are commonly prepared by replacing sensitive identity terms, i.e., the identity terms present in the instance are replaced with other identity terms that fall under the same sensitive category. Therefore, the efficacy of these methods depends heavily on the quality and comprehensiveness of identity pairs.  In this paper, we offer a two-step data augmentation process where (1) the former stage consists of a novel method for preparing a comprehensive list of identity pairs with word embeddings, and (2) the latter consists of leveraging prepared identity pairs list to enhance the training instances by applying three simple operations (namely identity pair replacement, identity term blindness, and identity pair swap).  We empirically show that the two-stage augmentation process leads to diverse identity pairs and an enhanced training set, with an improved counterfactual token-based fairness metric score on two well-known text classification tasks.
\end{abstract}


\maketitle

\section{Introduction}
Machine learning models learn the statistical patterns from historical instances to predict labels for future instances. These models are intended to learn bias; for example, a toxic text classifier is expected to understand the presence of toxic terms and assign a higher score to those terms compared to other non-toxic terms. However, models are not intended to discriminate between two instances based on sensitive identity terms, such as race or gender indication terms \cite{bolukbasi2016man, blodgett2017racial, webster2018mind}. \citet{dixon2018measuring} showed that due to the disproportionate distribution of terms in the training data, text classification models could unintentionally learn bias (defined as unintended bias) specific to the identity terms.

At the instance level, a related bias issue arises when a classifier assigns different scores to two nearly identical instances with different identity information.~\citet{garg2019counterfactual} study counterfactual fairness by considering the question: \textit{How would the prediction change if the sensitive identity attributes referenced in the text instance were different?} To address counterfactual fairness, it compares three approaches to address instance-level bias and introduces a counterfactual token fairness metric. Both measurement and mitigation methods discussed depend heavily on the counterfactual instances generation process. Counterfactual instances are generated by replacing identity terms present in the instance with other identity terms that fall in the same category. For example, replacing the term \textit{women} in the sentence \textit{"women should rule the world"} with the term \textit{men} to create a counterfactual sentence \textit{"men should rule the world"}. Counterfactual fairness methods, thus, demand information about identity pairs (such as women-men, gay-straight), and could extensively benefit from diverse identity pairs information and counterfactual generation mechanisms. However, \citet{garg2019counterfactual} limits the process to a manually prepared list of 50 identity pairs. 

\begin{figure}[h]
	\centering
	\includegraphics[width=0.45\textwidth]{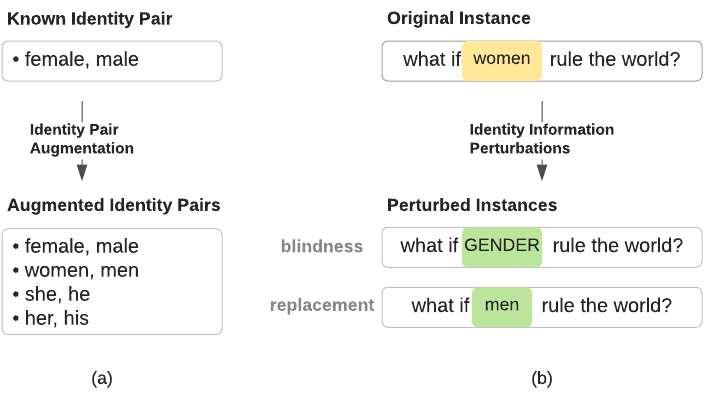}
	\caption{Overview of the IIDA process, (a) augments identity pairs for a known identity pair, (b) leverages augmented identity pairs for perturbing an example instance with blindness and replacement operations}
	\label{fig:process}
\end{figure}

In this paper, we present \textbf{I}dentity \textbf{I}nformation \textbf{D}ata \textbf{A}ugmentation methods, IIDA for brevity, to bolster the text classification models against unintended bias by automatically generating diverse counterfactual instances. Given a subset of known identity pairs \cite{borkan2019nuanced, garg2019counterfactual}, we first apply a simple but powerful method to augment the identity pairs with additional multifarious identity pairs that occur in the same context. We then use the expanded comprehensive identity pairs list to measure the counterfactual token fairness metric. Prepared identity pairs along with a set of instance perturbations operations like identity blindness, replacement, and swap, are used to produce counterfactual instances that are leveraged for training the model either in the form of augmentation or as a logit-pairing training scheme ~\cite{garg2019counterfactual, kannan2018adversarial}. In summary, the main contributions of our paper are: 
\begin{enumerate}
	\item \textbf{Identity Pair Augmentation (Section \ref{sec:ipg})}: We propose a novel word-embeddings based method to generate diverse identity pairs, as shown in Figure \ref{fig:process}a. Diverse identity pairs are leveraged for preparing counterfactual training instances and for measuring counterfactual token fairness \cite{garg2019counterfactual}.
	\item \textbf{Identity Information Perturbations (Section \ref{sec:iip})}: To enhance the training set with counterfactuals information,  we leverage augmented identity pairs to perform three types of text perturbation operations on training instances, as shown in Figure \ref{fig:process}b.
	\item \textbf{Effectiveness (Section \ref{sec:experiments})}: We methodically evaluate IIDA on two well-recognized unintended bias inducing text classification tasks, demonstrating that IIDA methods help mitigate the unintended bias while sustaining the model performance on both the tasks.
\end{enumerate}

\section{Related work}
\citet{garg2019counterfactual} propose a fairness metric, \textit{Counterfactual Token fairness} (CTF gap), for measuring model's bias where bias is computed by taking the absolute difference between model's prediction score on instance and its counterfactual, averaged over all records. Blindness, Counterfactual Augmentation, and Counterfactual Logit Pairing methods are also discussed by \cite{garg2019counterfactual} for optimizing counterfactual fairness while training text classification models. \citet{qian2019reducing} edit the loss function to equalize the probabilities of male and female words to alleviate bias in word-level language models. \citet{wei2019eda} presents data perturbation techniques for boosting performance of text classification models. Synonym replacement, random insertion/swap/deletion, perturbation operations are performed by \citet{wei2019eda}. Our work leverages the perturbation operations thought to bolster fairness metric.

\section{IIDA}
We propose a two-step data augmentation process where we first augment known identity pairs with diverse identity pairs to make it comprehensive. We then apply perturbation operations to enhance the training set and for the counterfactual token fairness metric. Table \ref{tab:knownidentitypairs} presents a sampled set of known identity pairs and other possible diverse pairs.

\begin{table}
	\begin{tabular}{ |p{1.2cm} | p{5.4cm}| }
		\hline
		\textbf{Known Pair} & \textbf{Other Possible Pairs} \\
		\hline
		man-woman & he-she, boy-girl, guy-wife, boy-person, him-girl, father-mother\\
		\hline
		blacks-whites & black-american, latino-mexican, hispanic-african\_american, mexican-african \\
		\hline
		trump-biden & trump-obama, palin-mitt, barack-clinton, palin-hillary, trump-hillary   \\
		\hline
	\end{tabular}
	\caption{Identity Pairs}\label{tab:knownidentitypairs}
\end{table}

\subsection{Identity Pair Augmentation}
\label{sec:ipg}
We present the identity pair augmentation procedure in Algorithm \ref{alg:algorithm}. Given a set of known identity pairs $I$, we leverage the word embedding model~\citep{mikolov2013efficient, pennington2014glove, bojanowski2017enriching}, to augment identity pairs. For each known identity pair $(i_1, i_2)$, we use embedding model $M$ to extract top $top_{k1}$ most similar words for identity terms $i_1$ and $i_2$, and form a cartesian product (line 3) of the two extracted words set denoted by $J$. $J$ captures the set of word pairs that possibly occur in the same context with respect to the input pair, and to filter out the pairs that don't occur in the same context (i.e., to remove noisy pairs), we perform the steps presented in lines 5-9. For each pair $(j_1, j_2)$, we extract top $top_{k2}$ most similar words for identity terms $j_1$ and $j_2$ and compute a similarity score (with sets similarity methods like Jaccard-index) between the two sets. We consider $(j_1, j_2)$ a valid identity pair only if the similarity score computed before is above a selected threshold value $\tau$. Augmented identity terms are simply a set of unique terms that are part of augmented identity pairs. We present the parameter values in the experiment section below and show that the proposed procedure works well in practice.

We also investigated the word embedding model with analogy operation where given input $term1$ and $term2$, we iterate over $top_K$ most similar terms/words for $term1$, and we represent each similar term by $term1'$. We then find analogy $term2'$  with the following setting:
\begin{equation*}
w(term2') = w(term2) + w(term1') - w(term1)
\end{equation*}
We considered $(term1', term2')$ as an augmented identity pair. We found that even a single generic term in $top_K$ terms could lead to invalid out-of-context identity pairs. For example - say we consider $(man, woman)$ as a known identity pair, then if term $`person`$ occurs in $top_{10}$ similar words to term $`man`$, finding analogy would lead to erroneous terms like $`individual`, `each`, `whether`, `every`$. Algorithm \ref{alg:algorithm} we propose avoids this by doing a similarity check over $top_{k2}$ similar terms and using a similarity threshold to avoid invalid pairs. We observed that in practice the identity pairs generated by Algorithm \ref{alg:algorithm} are diverse and valid in most cases. We share a sampled list of the identity pairs generated by Algorithm \ref{alg:algorithm} in Appendix \ref{apx:generatedlist} section.

\begin{algorithm}[tb]
	\caption{Identity Pair Augmentation}
	\label{alg:algorithm}
	\begin{flushleft}
		\textbf{Notations}: 
		\\ $\zeta$ -\  function to get top-k most similar words 
		\\ $\psi$ -\ function to get similarity score
		\\
		\textbf{Input}: 
		\\ $I$ -\ known identity pairs
		\\ $M$ -\ word embedding model
		\\ $\tau$ -\ similarity score threshold
		\\ $top_{k1}$ -\ top-k value for augmentation
		\\ $top_{k2}$ -\ top-k value for similarity check
		\\
		\textbf{Output}: $O$ -\ augmented identity pairs \\
		\textbf{Process:}
	\end{flushleft}
	\begin{algorithmic}[1] 
		\STATE $O$ := $set(I)$
		\STATE $\forall (i_1, i_2) \in I$
		\STATE \quad \quad $J$ := $\zeta(M, i_1, top_{k1}) \times \zeta(M, i_2, top_{k1})$
		\STATE \quad \quad $\forall (j_1, j_2) \in J$
		\STATE \quad \quad \quad \quad $s_1$ := $\zeta(M, j_1, top_{k2})$
		\STATE \quad \quad \quad \quad $s_2$ := $\zeta(M, j_2, top_{k2})$
		\STATE \quad \quad \quad \quad \textbf{if} $\psi(s_1, s_2) \ge \tau$ \textbf{do}
		\STATE \quad \quad \quad \quad \quad \quad $O.add((j_1, j_2))$
		\STATE \quad \quad \quad \quad \textbf{endif}
		\STATE \textbf{return} $O$
	\end{algorithmic}
\end{algorithm}

\subsection{Identity Information Perturbations}
\label{sec:iip}
We here present the details to leverage augmented identity terms and pairs for generating counterfactual instances. We additionally perform two other operations, blindness, and swap, for instance perturbation. On each instance in the training set, we perform the following operations:

\begin{enumerate}[label=\Alph*]
\item \textbf{Identity Pair Replacement (\textit{IPR})}: Randomly sample $n_r$ identity pairs from the augmented identity pairs and replace all the first term occurrences with the second term of the pair.
\item \textbf{Identity Term Blindness (\textit{ITB})}: Randomly sample $n_b$ identity terms from the augmented identity terms and replace all the occurrences of the sampled terms with the blindness placeholder. Each term could have a dedicated placeholder; for example, a gender representing term could have a placeholder  \textit{`GENDER\_TOKEN'}, and similarly, \textit{`NAME\_TOKEN'} for a public figure name.
\item \textbf{Identity Pair Swap (\textit{IPS})}: Randomly sample $n_s$ identity pairs from the augmented identity pairs and swap all the sampled pairs' occurrences.
\end{enumerate}

We selected these operations based on the research by \citet{wei2019eda}, and heuristics that counterfactual and swap instances help classifier learn and optimize diverse data while blindness operations help hide sensitive information specific to categories with several sub-categories (e.g., public figure names).
 \textit{IPR} operation on an input instance $x$, leads to the generation of diverse counterfactual instances denoted by $\phi_r(x)$. Similarly, \textit{ITB} operation leads to the generation of blindness instances represented by $\phi_b(x)$, and the collection of instances generated by \textit{IPS} operation are characterized by $\phi_s(x)$. \textit{IPR} and \textit{ITB} operations are also considered by \cite{garg2019counterfactual} but with diverse identity pairs it helps to mitigate bias, as we demonstrate in Section \ref{sec:experiments}.

We further leverage $\phi_r$ and $\phi_b$ to update the logit pairing training scheme proposed by \cite{garg2019counterfactual} to the following objective:
\begin{multline}
\sum_{x \in X} C(f(x), y) \\ 
+ \lambda_1\sum_{x\in X}\mathop{\mathbb{E}}_{x_r\sim Unif[\phi_r(x)]} {\mid g(x) - g(x_r)\mid} \\
+ \lambda_2\sum_{x\in X}\mathop{\mathbb{E}}_{x_b\sim Unif[\phi_b(x)]} {\mid g(x) - g(x_b)\mid}
\end{multline}
where $X$, $C$, $f$, $y$ denote the training set, cost function, classifier, and true prediction value respectively, $g$ produces a logit, and $\lambda_1$, $\lambda_2$ are penalty hyperparameters. Counterfactual and blinded instances individually have shown to mitigate bias \cite{garg2019counterfactual}, and we  experiment the logit pairing training scheme with a combination of both settings. Both settings act complementary, i.e., sensitive categories with limited list of identity pairs (for example, race sensitive category) could be handled by counterfactual instances, and categories with exhaustive list of identity pairs (for example, public figure names category)  could be handled by blinded instances. Swap instances $\phi_s$ could also be considered with a logit pairing scheme, but we keep it out of paper scope.

\section{Experiments}
\label{sec:experiments}
\subsection{Benchmark Datasets}
We perform experiments on two public text classification benchmark datasets: (1) Toxic Comment Classification Dataset \footnote[1]{https://www.kaggle.com/c/jigsaw-toxic-comment-classification-challenge}, (2) Hate Speech Dataset \footnote[2]{https://github.com/Vicomtech/hate-speech-dataset} \cite{de2018hate}. We combine six classes (toxic, severe\_toxic, obscene, threat, insult, identity\_hate) in the toxic comment classification dataset under positive label class `Toxic` and work with a binary classification setting. Dataset statistics are present in Appendix \ref{apx:experimentdetails}.

\subsection{Experimental Setting}
For identity pairs augmentation, we use 
GloVe 42B-tokens \footnote[3]{https://nlp.stanford.edu/projects/glove/} embedding model \cite{pennington2014glove} with Jaccard-index for computing similarity score between two word sets. We set $\tau=0.25$, $top_{k1}=10$, $top_{k2}=500$, and perform lemmatization operation on the generated identity pairs to remove similar form redundant pairs. For Hate Speech Dataset, we keep $n_r=50\%$, $n_s=10\%$ of the augmented identity pairs, and $n_b=10\%$ of the augmented identity terms, for identity replacement, swap, and blindness operations respectively. For Toxic Comment Classification Dataset, we keep $n_r=10\%$, $n_s=10\%$, and $n_b=10\%$.

We work with a set of $10$ known identity pairs (shown in Appendix Section \ref{apx:generatedlist}, Table \ref{tab:appendixidentitypairs}), and use the proposed Algorithm \ref{alg:algorithm} to augment identity pairs (shown in Appendix Section B, Table \ref{tab:appendixgenidentitypairs}).

We conduct experiments with the Convolutional Neural Network (CNN) classifier \cite{kim-2014-convolutional}. Previous studies on Toxic Comment Classification and Hate Speech Classification datasets have demonstrated good performance with the CNN classifier \cite{garg2019counterfactual}, and therefore we use it for experimentation. We use CNN with filters of sizes 3, 4 and 5 with each having 512 filters for all experiments. For the Hate Speech dataset, we train for 25 epochs with early stopping criteria. For the Toxic Comment Classification dataset, we train for 3 epochs. Training batch size across all experiments is 128. We use categorical cross entropy as a loss function. We use Adam with a learning rate $3\mathrm{e}{-4}$ as an optimizer. For CTF metric calculation, we generate all possible combinations of replacements possible with the set of augmented identity pairs. 

\subsection{Results}

We use the CTF gap metric proposed by \cite{garg2019counterfactual} as a fairness evaluation metric where the CTF gap, for an instance, is defined as:

\begin{equation}
CTF\,gap_{\phi_r(x)} = \mathop{\mathbb{E}}_{x_r\sim Unif[\phi_r(x)]} {\mid f(x) - f(x_r)\mid}
\end{equation}

CTF gap over test set is the mean of CTF gap over all individual instances (Equation 2) with all possible valid counterfactual instances. We form test set counterfactual instances with augmented identity pairs and \textit{IPR} operation. Proposed methods could also be tested with other perturbation analysis methods, like \citet{prabhakaran2019perturbation}, but we limit the paper's scope to CTF metric.

\begin{figure}[h]
	\centering
	\includegraphics[width=0.45\textwidth]{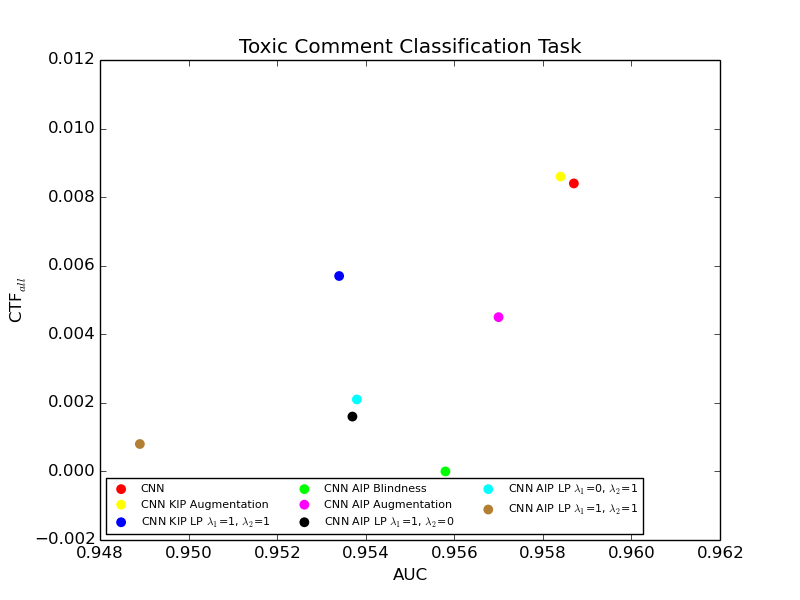}
	\caption{AUC vs. CTF\textsubscript{all} comparison on Toxic Comment Classification Task}
	\label{fig:aucvsctf1}
\end{figure}

\begin{figure}[h]
	\centering
	\includegraphics[width=0.45\textwidth]{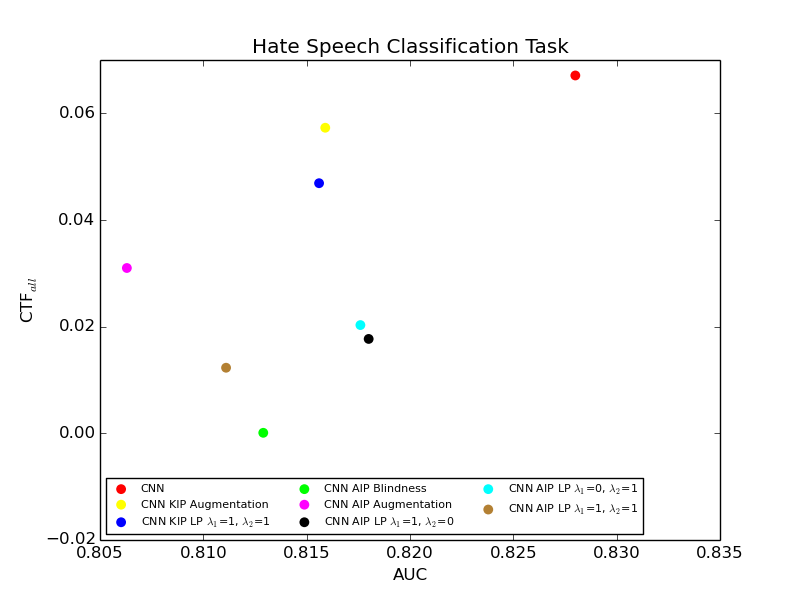}
	\caption{AUC vs. CTF\textsubscript{all} comparison on Hate Speech Classification Task}
	\label{fig:aucvsctf2}
\end{figure}

Table \ref{tab:jigsawresultsf} and \ref{tab:jigsawresultss} show performance metric and fairness metric results respectively, for Toxic Comment Classification task. We denote Known Identity Pairs set by \textit{`KIP'} and Augmented Identity Pairs set by \textit{`AIP'}. We measure Accuracy (at threshold = $0.5$) and CTF gap at the label class level (positive class denoted by \textit{`pos'}, negative class denoted by \textit{`neg'}, and combined denoted by \textit{`all'}) to better understand the impact of mitigation methods. Blindness method (CNN AIP Blindness) by design results in a CTF gap value close to 0. Blindness method, however, misses on a lot of information that may be useful for classification tasks, and hence, is not usually recommended for bias mitigation. Other shortcomings of the blindness method are discussed further in \citep{garg2019counterfactual}. The augmentation method (CNN AIP Augmentation) adds $\phi_r$, $\phi_b$, and $\phi_s$ instances for each training instance to the training set and trains the model with the augmented set. CNN AIP Augmentation method, as observed empirically, reduces the CTF gap value. The logit pairing method (CNN AIP LP), as expected, performs best with a significant reduction in the CTF gap value. We present the results with three sets of penalty values, i.e., (1) $\lambda_1=1$, $\lambda_2=0$, (2) $\lambda_1=0$, $\lambda_2=1$, (3) $\lambda_1=1$, $\lambda_2=1$. CNN AIP LP $\lambda_1=1$, $\lambda_2=1$ works better than other methods. Appendix C presents the details about other values of $\lambda_1, \lambda_2$. We also show with two classifier settings (namely CNN KIP Augmentation and CNN KIP LP) that the proposed identity pairs augmentation method leads to improved CTF gap value. Overall, all the mitigation methods reduce the CTF gap value without harming the classifier's overall Accuracy and AUC score. We observe similar results for Hate Speech Classification task, as shown in Table \ref{tab:hatespeechresultsf} and \ref{tab:hatespeechresultss},  where CNN AIP LP $\lambda_1=1$, $\lambda_2=1$ works best, i.e., shows low CTF gap value  and doesn't affect model's performance metrics. We visualize AUC vs. CTF\textsubscript{all} comparison in Figure \ref{fig:aucvsctf1} and  \ref{fig:aucvsctf2} to better understand the performance vs. fairness metric trade-off.

\begin{table}
	\centering
	\begin{tabular}{|*{5}{c|}}
		\hline
		\multirowcell{3}{\backslashbox{Model}{Metric}} & \multicolumn{3}{c|}{Accuracy} &
		\\
		\cline{2-4}
		& \makecell{\\ $Acc_{pos}$ \\} & \makecell{\\ $Acc_{neg}$ \\} & \makecell{\\$Acc_{all}$\\} & AUC \\ \hline \hline
		CNN & 0.8271 & 0.9264 & 0.9167 & 0.9587 \\ 
		\hline
		CNN KIP Augmentation & 0.8708	& 0.9050 & 0.9016 & 0.9584 \\ 
		CNN KIP LP $\lambda_1=1, \lambda_2=1$ & 0.7533 & 0.9468 & 0.9278 & 0.9534 \\ \hline
		CNN AIP Blindness & 0.8258 & 0.9217	& 0.9123 & 0.9558\\
		CNN AIP Augmentation & 0.8520 & 0.9139 & 0.9078 & 0.9570 \\
		CNN AIP LP $\lambda_1=1, \lambda_2=0$ & 0.7804 & 0.9364	& 0.9211 & 0.9537 \\
		CNN AIP LP $\lambda_1=0, \lambda_2=1$ & 0.8117 & 0.9246	& 0.9135 & 0.9538 \\
		CNN AIP LP $\lambda_1=1, \lambda_2=1$ & 0.7327 & 0.9460 & 0.9251 & 0.9489 \\ \hline
	\end{tabular}
	\caption{Toxic Comment Classification Task : Performance Metric Results}
	\label{tab:jigsawresultsf}
\end{table}

\begin{table}
	\centering
	\begin{tabular}{|*{4}{c|}}
		\hline
		\multirowcell{3}{\backslashbox{Model}{Metric}} & \multicolumn{3}{c|}{CTF Gap}
		\\
		\cline{2-4}
		& \makecell{\\ $CTF_{pos}$ \\} & \makecell{\\ $CTF_{neg}$ \\} & \makecell{\\$CTF_{all}$\\} \\ \hline \hline
		CNN & 0.0229 & 0.0069 & 0.0084  \\ 
		\hline
		CNN KIP Augmentation & 0.0198 & 0.0074	& 0.0086  \\ 
		CNN KIP LP $\lambda_1=1, \lambda_2=1$ &	0.0205 & 0.0042	& 0.0057  \\ \hline
		CNN AIP Blindness & 0.0000 & 0.0000 & 0.0000 \\
		CNN AIP Augmentation & 0.0119	& 0.0037 & 0.0045  \\
		CNN AIP LP $\lambda_1=1, \lambda_2=0$ & 0.0059 & 0.0011	& 0.0016  \\
		CNN AIP LP $\lambda_1=0, \lambda_2=1$ & 0.0064	& 0.0017 & 0.0021 \\
		CNN AIP LP $\lambda_1=1, \lambda_2=1$ & \textbf{0.0027}	& \textbf{0.0007} & \textbf{0.0008} \\ \hline
	\end{tabular}
	\caption{Toxic Comment Classification Task : Fairness Metric Results}
	\label{tab:jigsawresultss}
\end{table}

\begin{table}
	\centering
	\begin{tabular}{|*{5}{c|}}
		\hline
		\multirowcell{3}{\backslashbox{Model}{Metric}} & \multicolumn{3}{c|}{Accuracy} &
		\\
		\cline{2-4}
		& \makecell{\\ $Acc_{pos}$ \\} & \makecell{\\ $Acc_{neg}$ \\} & \makecell{\\$Acc_{all}$\\} & AUC \\ \hline \hline
		CNN & 0.8368 & 0.6653	& 0.7510 & 0.8280 \\ \hline
		CNN KIP Augmentation & 0.7657	& 0.7406 & 0.7531 & 0.8159 \\ 
		CNN KIP LP $\lambda_1=1, \lambda_2=1$ & 0.7406 & 0.7155 & 0.7280 & 0.8156 \\ \hline
		CNN AIP Blindness & 0.7322 & 0.7490 & 0.7406 & 0.8129 \\
		CNN AIP Augmentation & 0.8075 & 0.6569 & 0.7322	& 0.8063 \\
		CNN AIP LP $\lambda_1=1, \lambda_2=0$ & 0.8159 & 0.6778 & 0.7469 & 0.8180 \\
		CNN AIP LP $\lambda_1=0, \lambda_2=1$ & 0.8075 & 0.6987	& 0.7531 & 0.8176 \\
		CNN AIP LP $\lambda_1=1, \lambda_2=1$ & 0.7874	& 0.6769 & 0.7322 & 0.8111 \\ \hline
	\end{tabular}
	\caption{Hate Speech Classification Task : Performance Metric Results}
	\label{tab:hatespeechresultsf}
\end{table}

\begin{table}
	\centering
	\begin{tabular}{|*{4}{c|}}
		\hline
		\multirowcell{3}{\backslashbox{Model}{Metric}} & \multicolumn{3}{c|}{CTF Gap}
		\\
		\cline{2-4}
		& \makecell{\\ $CTF_{pos}$ \\} & \makecell{\\ $CTF_{neg}$ \\} & \makecell{\\$CTF_{all}$\\} \\ \hline \hline
		CNN & 0.0598 & 0.0776 & 0.0671 \\ \hline
		CNN KIP Augmentation & 0.0507 & 0.0668 &	0.0573 \\ 
		CNN KIP LP $\lambda_1=1, \lambda_2=1$ &	0.0428 & 0.0529	& 0.0469  \\ \hline
		CNN AIP Blindness & 0.0001 & 0.0000 & 0.0001 \\
		CNN AIP Augmentation & 0.0342 & 0.0264 & 0.0310	 \\
		CNN AIP LP $\lambda_1=1, \lambda_2=0$ & 0.0139 & 0.0230 & 0.0177  \\
		CNN AIP LP $\lambda_1=0, \lambda_2=1$ &	0.0170 & 0.0250 & 0.0203  \\
		CNN AIP LP $\lambda_1=1, \lambda_2=1$ & \textbf{0.0110}	& \textbf{0.0141} & \textbf{0.0123} \\ \hline
	\end{tabular}
	\caption{Hate Speech Classification Task : Fairness Metric Results}
	\label{tab:hatespeechresultss}
\end{table}

\section{Conclusion}
We demonstrate the efficacy of proposed Identity Information Data Augmentation methods, IIDA, to help mitigate bias in text classification tasks. The two-step augmentation setup we propose leads to diverse identity pairs and enhanced training instances based on simple operations like identity replacement, blindness, and swap. We further show that amalgam of counterfactual and blindness based logit-pairing training scheme results in a more fair classifier. We believe that IIDA's simple methods could be efficiently leveraged in practice. Continued work on this topic could examine identity pairs validation, perform analysis at different subgroups level, study the effects of out-of-vocabulary tokens, explore perturbation operations enhancement to create and validate meaningful instances.

\bibliographystyle{ACM-Reference-Format}
\bibliography{sample-base}


\begin{thebibliography}{15}


\ifx \showCODEN    \undefined \def \showCODEN     #1{\unskip}     \fi
\ifx \showDOI      \undefined \def \showDOI       #1{#1}\fi
\ifx \showISBNx    \undefined \def \showISBNx     #1{\unskip}     \fi
\ifx \showISBNxiii \undefined \def \showISBNxiii  #1{\unskip}     \fi
\ifx \showISSN     \undefined \def \showISSN      #1{\unskip}     \fi
\ifx \showLCCN     \undefined \def \showLCCN      #1{\unskip}     \fi
\ifx \shownote     \undefined \def \shownote      #1{#1}          \fi
\ifx \showarticletitle \undefined \def \showarticletitle #1{#1}   \fi
\ifx \showURL      \undefined \def \showURL       {\relax}        \fi
\providecommand\bibfield[2]{#2}
\providecommand\bibinfo[2]{#2}
\providecommand\natexlab[1]{#1}
\providecommand\showeprint[2][]{arXiv:#2}

\bibitem[\protect\citeauthoryear{Blodgett and O'Connor}{Blodgett and
  O'Connor}{2017}]%
        {blodgett2017racial}
\bibfield{author}{\bibinfo{person}{Su~Lin Blodgett} {and}
  \bibinfo{person}{Brendan O'Connor}.} \bibinfo{year}{2017}\natexlab{}.
\newblock \showarticletitle{Racial disparity in natural language processing: A
  case study of social media african-american english}.
\newblock \bibinfo{journal}{\emph{Talk at the 2017 Workshop on Fairness,
  Accountability, and Transparency in Machine Learning}}
  (\bibinfo{year}{2017}).
\newblock


\bibitem[\protect\citeauthoryear{Bojanowski, Grave, Joulin, and
  Mikolov}{Bojanowski et~al\mbox{.}}{2017}]%
        {bojanowski2017enriching}
\bibfield{author}{\bibinfo{person}{Piotr Bojanowski}, \bibinfo{person}{Edouard
  Grave}, \bibinfo{person}{Armand Joulin}, {and} \bibinfo{person}{Tomas
  Mikolov}.} \bibinfo{year}{2017}\natexlab{}.
\newblock \showarticletitle{Enriching word vectors with subword information}.
\newblock \bibinfo{journal}{\emph{Transactions of the Association for
  Computational Linguistics}}  \bibinfo{volume}{5} (\bibinfo{year}{2017}),
  \bibinfo{pages}{135--146}.
\newblock


\bibitem[\protect\citeauthoryear{Bolukbasi, Chang, Zou, Saligrama, and
  Kalai}{Bolukbasi et~al\mbox{.}}{2016}]%
        {bolukbasi2016man}
\bibfield{author}{\bibinfo{person}{Tolga Bolukbasi}, \bibinfo{person}{Kai-Wei
  Chang}, \bibinfo{person}{James Zou}, \bibinfo{person}{Venkatesh Saligrama},
  {and} \bibinfo{person}{Adam Kalai}.} \bibinfo{year}{2016}\natexlab{}.
\newblock \showarticletitle{Man is to computer programmer as woman is to
  homemaker? debiasing word embeddings}.
\newblock \bibinfo{journal}{\emph{NeurIPS}} (\bibinfo{year}{2016}).
\newblock


\bibitem[\protect\citeauthoryear{Borkan, Dixon, Sorensen, Thain, and
  Vasserman}{Borkan et~al\mbox{.}}{2019}]%
        {borkan2019nuanced}
\bibfield{author}{\bibinfo{person}{Daniel Borkan}, \bibinfo{person}{Lucas
  Dixon}, \bibinfo{person}{Jeffrey Sorensen}, \bibinfo{person}{Nithum Thain},
  {and} \bibinfo{person}{Lucy Vasserman}.} \bibinfo{year}{2019}\natexlab{}.
\newblock \showarticletitle{Nuanced metrics for measuring unintended bias with
  real data for text classification}. In \bibinfo{booktitle}{\emph{Companion
  Proceedings of The 2019 World Wide Web Conference}}.
  \bibinfo{pages}{491--500}.
\newblock


\bibitem[\protect\citeauthoryear{de~Gibert, Perez, Garc{\'\i}a-Pablos, and
  Cuadros}{de~Gibert et~al\mbox{.}}{2018}]%
        {de2018hate}
\bibfield{author}{\bibinfo{person}{Ona de Gibert}, \bibinfo{person}{Naiara
  Perez}, \bibinfo{person}{Aitor Garc{\'\i}a-Pablos}, {and}
  \bibinfo{person}{Montse Cuadros}.} \bibinfo{year}{2018}\natexlab{}.
\newblock \showarticletitle{Hate speech dataset from a white supremacy forum}.
\newblock \bibinfo{journal}{\emph{Proceedings of the 2nd Workshop on Abusive
  Language Online ({ALW}2)}} (\bibinfo{year}{2018}).
\newblock


\bibitem[\protect\citeauthoryear{Dixon, Li, Sorensen, Thain, and
  Vasserman}{Dixon et~al\mbox{.}}{2018}]%
        {dixon2018measuring}
\bibfield{author}{\bibinfo{person}{Lucas Dixon}, \bibinfo{person}{John Li},
  \bibinfo{person}{Jeffrey Sorensen}, \bibinfo{person}{Nithum Thain}, {and}
  \bibinfo{person}{Lucy Vasserman}.} \bibinfo{year}{2018}\natexlab{}.
\newblock \showarticletitle{Measuring and mitigating unintended bias in text
  classification}. In \bibinfo{booktitle}{\emph{Proceedings of the 2018
  AAAI/ACM Conference on AI, Ethics, and Society}}. \bibinfo{pages}{67--73}.
\newblock


\bibitem[\protect\citeauthoryear{Garg, Perot, Limtiaco, Taly, Chi, and
  Beutel}{Garg et~al\mbox{.}}{2019}]%
        {garg2019counterfactual}
\bibfield{author}{\bibinfo{person}{Sahaj Garg}, \bibinfo{person}{Vincent
  Perot}, \bibinfo{person}{Nicole Limtiaco}, \bibinfo{person}{Ankur Taly},
  \bibinfo{person}{Ed~H Chi}, {and} \bibinfo{person}{Alex Beutel}.}
  \bibinfo{year}{2019}\natexlab{}.
\newblock \showarticletitle{Counterfactual fairness in text classification
  through robustness}. In \bibinfo{booktitle}{\emph{Proceedings of the 2019
  AAAI/ACM Conference on AI, Ethics, and Society}}. \bibinfo{pages}{219--226}.
\newblock


\bibitem[\protect\citeauthoryear{Kannan, Kurakin, and Goodfellow}{Kannan
  et~al\mbox{.}}{2018}]%
        {kannan2018adversarial}
\bibfield{author}{\bibinfo{person}{Harini Kannan}, \bibinfo{person}{Alexey
  Kurakin}, {and} \bibinfo{person}{Ian Goodfellow}.}
  \bibinfo{year}{2018}\natexlab{}.
\newblock \showarticletitle{Adversarial logit pairing}.
\newblock \bibinfo{journal}{\emph{CoRR abs/1803.06373}} (\bibinfo{year}{2018}).
\newblock


\bibitem[\protect\citeauthoryear{Kim}{Kim}{2014}]%
        {kim-2014-convolutional}
\bibfield{author}{\bibinfo{person}{Yoon Kim}.} \bibinfo{year}{2014}\natexlab{}.
\newblock \showarticletitle{Convolutional Neural Networks for Sentence
  Classification}. In \bibinfo{booktitle}{\emph{Proceedings of the 2014
  Conference on Empirical Methods in Natural Language Processing ({EMNLP})}}.
  \bibinfo{publisher}{Association for Computational Linguistics},
  \bibinfo{address}{Doha, Qatar}, \bibinfo{pages}{1746--1751}.
\newblock
\urldef\tempurl%
\url{https://doi.org/10.3115/v1/D14-1181}
\showDOI{\tempurl}


\bibitem[\protect\citeauthoryear{Mikolov, Chen, Corrado, and Dean}{Mikolov
  et~al\mbox{.}}{2013}]%
        {mikolov2013efficient}
\bibfield{author}{\bibinfo{person}{Tomas Mikolov}, \bibinfo{person}{Kai Chen},
  \bibinfo{person}{Greg Corrado}, {and} \bibinfo{person}{Jeffrey Dean}.}
  \bibinfo{year}{2013}\natexlab{}.
\newblock \showarticletitle{Efficient estimation of word representations in
  vector space}.
\newblock \bibinfo{journal}{\emph{International Conference on Learning
  Representations}} (\bibinfo{year}{2013}).
\newblock


\bibitem[\protect\citeauthoryear{Pennington, Socher, and Manning}{Pennington
  et~al\mbox{.}}{2014}]%
        {pennington2014glove}
\bibfield{author}{\bibinfo{person}{Jeffrey Pennington},
  \bibinfo{person}{Richard Socher}, {and} \bibinfo{person}{Christopher~D
  Manning}.} \bibinfo{year}{2014}\natexlab{}.
\newblock \showarticletitle{Glove: Global vectors for word representation}. In
  \bibinfo{booktitle}{\emph{Proceedings of the 2014 conference on empirical
  methods in natural language processing (EMNLP)}}.
  \bibinfo{pages}{1532--1543}.
\newblock


\bibitem[\protect\citeauthoryear{Prabhakaran, Hutchinson, and
  Mitchell}{Prabhakaran et~al\mbox{.}}{2019}]%
        {prabhakaran2019perturbation}
\bibfield{author}{\bibinfo{person}{Vinodkumar Prabhakaran},
  \bibinfo{person}{Ben Hutchinson}, {and} \bibinfo{person}{Margaret Mitchell}.}
  \bibinfo{year}{2019}\natexlab{}.
\newblock \showarticletitle{Perturbation sensitivity analysis to detect
  unintended model biases}.
\newblock \bibinfo{journal}{\emph{Proceedings of the Conference on Empirical
  Methods in Natural Language Processing (EMNLP)}} (\bibinfo{year}{2019}).
\newblock


\bibitem[\protect\citeauthoryear{Qian, Muaz, Zhang, and Hyun}{Qian
  et~al\mbox{.}}{2019}]%
        {qian2019reducing}
\bibfield{author}{\bibinfo{person}{Yusu Qian}, \bibinfo{person}{Urwa Muaz},
  \bibinfo{person}{Ben Zhang}, {and} \bibinfo{person}{Jae~Won Hyun}.}
  \bibinfo{year}{2019}\natexlab{}.
\newblock \showarticletitle{Reducing gender bias in word-level language models
  with a gender-equalizing loss function}.
\newblock \bibinfo{journal}{\emph{arXiv preprint arXiv:1905.12801}}
  (\bibinfo{year}{2019}).
\newblock


\bibitem[\protect\citeauthoryear{Webster, Recasens, Axelrod, and
  Baldridge}{Webster et~al\mbox{.}}{2018}]%
        {webster2018mind}
\bibfield{author}{\bibinfo{person}{Kellie Webster}, \bibinfo{person}{Marta
  Recasens}, \bibinfo{person}{Vera Axelrod}, {and} \bibinfo{person}{Jason
  Baldridge}.} \bibinfo{year}{2018}\natexlab{}.
\newblock \showarticletitle{Mind the gap: A balanced corpus of gendered
  ambiguous pronouns}.
\newblock \bibinfo{journal}{\emph{Transactions of the Association for
  Computational Linguistics}}  \bibinfo{volume}{6} (\bibinfo{year}{2018}),
  \bibinfo{pages}{605--617}.
\newblock


\bibitem[\protect\citeauthoryear{Wei and Zou}{Wei and Zou}{2019}]%
        {wei2019eda}
\bibfield{author}{\bibinfo{person}{Jason Wei} {and} \bibinfo{person}{Kai Zou}.}
  \bibinfo{year}{2019}\natexlab{}.
\newblock \showarticletitle{Eda: Easy data augmentation techniques for boosting
  performance on text classification tasks}.
\newblock \bibinfo{journal}{\emph{Proceedings of the 2019 Conference on
  Empirical Methods in Natural Language Processing and the 9th International
  Joint Conference on Natural Language Processing (EMNLP-IJCNLP)}}
  (\bibinfo{year}{2019}).
\newblock


\end{thebibliography}

\newpage
\appendix

\section{Experiment Details}
\label{apx:experimentdetails}
Dataset statistics are present in Table \ref{tab:jigsawDataset} and \ref{tab:hateSpeechDataset}.
\begin{table}[htb]
	\centering
	\begin{tabular}{|c|c|c|c|}
		\hline
		& \textbf{Toxic} & \textbf{Non-toxic} & \textbf{Total}\\
		\hline
		Train & 16225 & 143346 & 159571 \\
		Test & 6243 & 57735 & 63978 \\
		\hline
	\end{tabular}
	\caption{Toxic Comment Classification Dataset}\label{tab:jigsawDataset}
	\vspace{-10mm}
\end{table}

\begin{table}[htb]
	\centering
	\begin{tabular}{|c|c|c|c|}
		\hline
		& \textbf{Hate} & \textbf{No Hate} & \textbf{Total}\\
		\hline
		Train & 957 & 957 & 1914 \\
		Test & 239 & 239 & 478 \\
		\hline
	\end{tabular}
	\caption{Hate Speech Dataset}\label{tab:hateSpeechDataset}
	\vspace{-6mm}
\end{table}

We present results on the Hate Speech Dataset for other configurations of $\lambda_1, \lambda_2$ in Table \ref{tab:appendixhatespeechresultsf} and \ref{tab:appendixhatespeechresultss}.
\begin{table}[H]
	\centering
	\begin{tabular}{|*{5}{c|}}
		\hline
		\multirowcell{3}{\backslashbox{Model}{Metric}} & \multicolumn{3}{c|}{Accuracy} &
		\\
		\cline{2-4}
		& \makecell{\\ $Acc_{pos}$ \\} & \makecell{\\ $Acc_{neg}$ \\} & \makecell{\\$Acc_{all}$\\} & AUC \\ \hline \hline
		CNN AIP LP $\lambda_1=0.5, \lambda_2=0.5$ & 0.8243 & 0.6611	& 0.7427 & 0.8120 \\
		CNN AIP LP $\lambda_1=5, \lambda_2=5$ & 0.6946 & 0.5105 & 0.6025 & 0.6536 \\ \hline
	\end{tabular}
	\caption{Hate Speech Classification Task : Performance Metric Results}
	\label{tab:appendixhatespeechresultsf}
\end{table}

\begin{table}[H]
	\centering
	\begin{tabular}{|*{4}{c|}}
		\hline
		\multirowcell{3}{\backslashbox{Model}{Metric}} & \multicolumn{3}{c|}{CTF Gap}
		\\
		\cline{2-4}
		& \makecell{\\ $CTF_{pos}$ \\} & \makecell{\\ $CTF_{neg}$ \\} & \makecell{\\$CTF_{all}$\\} \\ \hline \hline
		CNN AIP LP $\lambda_1=0.5, \lambda_2=0.5$ &	0.0151 & 0.0219 & 0.0179  \\
		CNN AIP LP $\lambda_1=5, \lambda_2=5$ & 0.0006 & 0.0007	& 0.0006 \\ \hline
	\end{tabular}
	\caption{Hate Speech Classification Task : Fairness Metric Results}
	\label{tab:appendixhatespeechresultss}
\end{table}

\section{Generated Identity Pairs Set}
\label{apx:generatedlist}
\begin{table}[H]
	\centering
	\begin{tabular}{c|c}
		\hline
		\textbf{Term 1} & \textbf{Term 2}\\
		\hline
		man & woman \\
		democrat & republican \\
		trump & biden \\
		blacks & whites \\
		asian & latino \\
		china & america \\
		africa & europe \\
		christian & jewish \\
		buddhist & atheist \\
		homosexual & heterosexual \\
		\hline
	\end{tabular}
	\caption{Known Identity Pairs}\label{tab:appendixidentitypairs}
	\vspace{-4mm}
\end{table}

We here present a small sample of pairs generated by Algorithm 1.

\begin{table}[H]
	\centering
	\begin{tabular}{c|c}
		\hline
		\textbf{Term 1} & \textbf{Term 2}\\
		\hline
		female & male \\
		religious & christianity \\
		mexican & hispanic \\
		she & men \\
		japanese & asian \\
		woman & man \\
		she & he \\
		muslim & christian \\
		asian & latino \\
		barack & mitt \\
		gingrich & hillary \\
		american & african-american \\
		woman & him \\
		liberal & conservative \\
		american & black \\
		girl & boy \\
		lady & guy \\
		gay & lesbian \\
		jew & christian \\
		atheist & buddhist \\
		asia & australia \\
		america & japan \\
		latino & african-american \\
		she & person \\
		homosexual & homosexuality \\
		lady & boy \\
		she & him \\
		africa & zimbabwe \\
		australia & italy \\
		australia & america \\
		korea & europe \\
		\hline
	\end{tabular}
	\caption{Generated Identity Pairs}\label{tab:appendixgenidentitypairs}
\end{table}

\end{document}